\documentclass[letterpaper]{article}
\usepackage{uai2020}
\usepackage[margin=1in]{geometry}

\usepackage{times}

\title{Semi-supervised Sequential Generative Models}

\author{ \textbf{Michael Teng}$^1$\thanks{Correspondence to \texttt{mteng@robots.ox.ac.uk}},~~~\textbf{Tuan Anh Le}$^2$,~~~\textbf{Adam Scibior}$^3$,~~~\textbf{Frank Wood}$^{3,4}$ \\
$^1$ Department of Engineering Science, University of Oxford \\
$^2$ Department of Brain and Cognitive Sciences, MIT \\
$^3$ Department of Computer Science, University of British Columbia\\
$^4$ Montr\'eal Institute for Learning Algorithms (MILA)\\
}

\usepackage[T1]{fontenc}    
\usepackage{hyperref}       
\usepackage{url}            
\usepackage{booktabs}       
\usepackage{amsfonts}       
\usepackage{nicefrac}       
\usepackage{microtype}      
\usepackage{graphicx}
\usepackage{tikz}
\usetikzlibrary{fit,positioning}
\usetikzlibrary{positioning, fit, arrows.meta, shapes}
\usepackage{latexsym}
\usepackage{amssymb}
\usepackage{amsmath}
\usepackage{subcaption}
\usepackage{tabu}
\usepackage{adjustbox}
\usepackage{wrapfig}

\usepackage{amsmath}[2000/07/18] 
\usepackage{amssymb}[2002/01/22] 
\usepackage{amsfonts}
\usepackage{bm}
\usepackage{xfrac}
\usepackage{stmaryrd}
\usepackage{mathtools}
\usepackage{eucal}
\usepackage{multicol}
\usepackage{algorithm}
\usepackage{algorithmic}
\usepackage[parfill]{parskip}
\usepackage{xcolor}

\newcommand{\given}{\lvert}
\newcommand{\E}{\mathbb{E}}

\newcommand{\KL}{\textsc{kl}}

\DeclarePairedDelimiterX{\infdivx}[2]{(}{)}{%
	#1\;\delimsize\|\;#2%
}

\newcommand{\kl}{\textrm{KL}\infdivx}

\usepackage{mathtools}
\usepackage{amssymb}
\usepackage{setspace}
\usepackage[acronym,smallcaps,nowarn,section,nogroupskip,nonumberlist]{glossaries}

\usepackage{placeins}

\glsdisablehyper{}
\newacronym{ELBO}{elbo}{evidence lower bound}
\newacronym{VAE}{vae}{variational autoencoder}
\newacronym{IWAE}{iwae}{importance weighted autoencoder}
\newacronym{KL}{kl}{Kullback-Leibler}
\newacronym{SGD}{sgd}{stochastic gradient descent}
\newacronym{VIMCO}{vimco}{variational inference for Monte Carlo objectives}
\newacronym{WW}{ww}{wake-wake}
\newacronym{REINFORCE}{reinforce}{Reinforce}
\newacronym{LSTM}{lstm}{long short-term memory}

\usepackage[round]{natbib}

\begin{document}

\maketitle

\begin{abstract}

We introduce a novel objective for training deep generative time-series models with discrete latent variables for which supervision is only sparsely available. This instance of semi-supervised learning is challenging for existing methods, because the exponential number of possible discrete latent configurations results in high variance gradient estimators. We first overcome this problem by extending the standard semi-supervised generative modeling objective with reweighted wake-sleep. However, we find that this approach still suffers when the frequency of available labels varies between training sequences. Finally, we introduce a unified objective inspired by teacher-forcing and show that this approach is robust to variable length supervision. We call the resulting method caffeinated wake-sleep (CWS) to emphasize its additional dependence on real data. We demonstrate its effectiveness with experiments on MNIST, handwriting, and fruit fly trajectory data.
\end{abstract}

\section{INTRODUCTION}

In recent years there has been an explosion of interest in deep generative
models (DGM), which use neural networks transforming random inputs to learn complex
probability distributions. We particularly focus on the variational auto-encoder (VAE ;  \cite{kingma2013auto}) family of models, where the generative model is learned simultaneously with an associated inference network. This approach has been extended to settings with partially observed discrete variables \citep{kingma2014semi} and sequential data \citep{chung2015recurrent} but so far little work has been done on combining the two. In this paper we address this gap.

In many scenarios it is natural to assign discrete labels to specific intervals of time-series data and try to infer these labels from observations. For example, sequences of stock prices can be identified as periods of bear or bull markets, fragments of home CCTV footage can be classified as burglaries or other events, and heart rate signals can be used to classify cardiac health. In addition to the latent classification of observations in these applications, we may also wish to conditionally generate new sequences or predict ahead given some input sequence. Existing approaches to generative modeling in this context either require full supervision in the label space, and therefore are not able to leverage large amounts of unlabeled data, or are restricted to classification and can not generate new data \citep{chen2013dtw, wei2006semi}.

Additionally, many methods for semi-supervised learning of static DGM tasks define separate unsupervised and supervised loss terms in the training objective, which are weighted to reflect the overall supervision rate in the dataset \citep{kingma2014semi}. Though this class of approaches is extendable to the time-series setting, the optimization of model and inference network parameters become unstable when there is an uneven contribution of partial labels to the supervised term, as a result of varying supervision rates in each training example. Because of this, the state-of-the-art approaches to semi-supervised learning, including Virtual Adversarial Training \citep{miyato2018virtual}, Mean Teacher \citep{tarvainen2017mean}, and entropy minimization \citep{grandvalet2005semi}, cannot be easily adapted to modeling time-series data.

In this work we derive two novel objectives for training DGMs with discrete latent variables in a semi-supervised fashion. They are both based on the reweighted wake-sleep (RWS) algorithm
\citep{bornschein2014reweighted}, which avoids using high variance score function estimators for expectations over the discrete latents. The first one, which we call semi-supervised wake-sleep (SSWS), is obtained by extending RWS with a supervised classification term. While it performs well when labels available in the training set are regularly distributed per training example, it suffers from optimization problems when they are not, which can be unavoidable in real world datasets. To overcome this problem, we introduce a new approach which performs wake-sleep style optimization using a single objective that incorporates both supervised and unsupervised terms for every gradient update. We call this approach caffeinated wake-sleep (CWS).

Although we are explicitly targeting the sequential setting, our objectives can also be useful in non-sequential settings, especially when the discrete latent space is too large to be fully enumerated. We evaluate our objectives on MNIST, a handwriting dataset, and a dataset of fruit fly trajectories labeled with behavior classes. In the rest of the paper, Section 2 reviews the relevant background information,
Section 3 describes our model and training objective, and Section 4 presents the experimental results.

\section{BACKGROUND}

\subsection{VARIATIONAL AUTO-ENCODERS}

Variational auto-encoders \citep{kingma2013auto, rezende2014stochastic} are an approach to deep generative modeling
where we simultaneously learn a generative model $p_\theta$ and an amortized
inference network $q_\phi$, both parameterized by neural networks. For any given observation $\mathbf x$, the inference network variationally approximates the intractable posterior distribution $p_\theta(\mathbf z|\mathbf x)$ over a latent variable $\mathbf z$ as $q_\phi(\mathbf z|\mathbf x)$.
In order to train both networks simultaneously, \citet{kingma2013auto} propose maximizing the \gls{ELBO}, which is a sum over ELBOs for individual data points defined as
\begin{align}
  \mathcal{L}(\theta, \phi, \mathbf x) :&= \log p_\theta(\mathbf x) - \kl{q_\phi(\mathbf z | \mathbf x)}{p_\theta(\mathbf z | \mathbf x)}\nonumber \\
  &= \mathbb E_{q_\phi(\mathbf z | \mathbf x)}\left[\log \frac{p_\theta(\mathbf z, \mathbf x)}{q_\phi(\mathbf z | \mathbf x)}\right] \label{eq:elbo-formula}
\end{align}
The ELBO approximates log-marginal likelihood $\log p_\theta(\mathbf x)$, making it a good target objective for learning $\theta$, and is proportional to negative $\mathrm{KL}(q_\phi|p_\theta)$, making it a good target objective for learning $\phi$.

\cite{burda2015importance} propose an extension to the \gls{VAE} where, for a given number of samples $K$, the single-datapoint objective is instead defined as
\begin{equation}\small{
    \mathcal{L}_{\text{}}^K(\theta, \phi, \mathbf x) = \mathbb E_{\mathbf z_{1:K} \sim  q_\phi}\left[\log\left(\frac{1}{K} \sum_{k = 1}^K \frac{p_\theta(\mathbf z_k, \mathbf x)}{q_\phi(\mathbf z_k | \mathbf x)}\right)\right] \label{eq:iwae-formula}
    }
\end{equation}
This yields a tighter bound for $\log p_\theta(\mathbf x)$, which is desirable for learning $\theta$.

In either case, the estimator of the gradient with respect to
$\phi$ is typically high variance if $\mathbf z$ includes any discrete variables and therefore not suitable for gradient-based optimization.
Various authors have proposed to reduce this variance using continuous
relaxation \citet{maddison2016concrete,jang2016categorical} or control-variate
methods \citep{mnih2016variational, mnih2014neural, tucker2017rebar, grathwohl2017backpropagation} with varying degrees of success.

An alternative approach is to use the reweighted wake-sleep (RWS) algorithm \citep{bornschein2014reweighted} which uses separate objectives for $\theta$ and $\phi$, interleaving the corresponding gradient steps. The target for $\theta$ is $\mathcal{L}^K_{\text{IWAE}}(\theta,\phi,\mathbf x)$, while the target for $\phi$ is $-\mathrm{KL}(p_\theta || q_\phi)$. In the latter the expectation with respect to $p_\theta$ is approximated by importance sampling from $q_\phi$. \cite{le2018revisiting} show that using two separate objectives avoids the problems with high variance gradient estimates in the presence of discrete latent variables and avoids learning problems discussed by \cite{rainforth2018tighter}. \cite{ba2015learning} demonstrate RWS as a viable method for time-series modeling by using it to train recurrent attention models.

\subsection{SEMI-SUPERVISED VAE}\label{sec:kingma}

In certain situations it is desirable to extend the VAE with an interpretable
latent variable $y$, the canonical example being learning on the MNIST dataset
where $y$ is the digit label, $\mathbf z$ is ``style,'' and $\mathbf x$ is the image. \cite{kingma2014semi} consider the setting where the label $y$ is only scarcely available and the dataset consists of many instances of $\mathbf x$ and relatively few
instances of $(\mathbf x,y)$, which is a typical semi-supervised learning setting. They
choose a factorization $p_\theta(\mathbf x,y,\mathbf z) = p_\theta(\mathbf x | y,\mathbf z) p_\theta(y)
p_\theta(\mathbf z)$ and $q_\phi(y,\mathbf z | \mathbf x) = q_\phi(\mathbf z | \mathbf x,y) q_\phi(y | \mathbf x)$. A naive
optimization objective can be constructed by summing standard ELBOs for $\mathbf x$ and
$(\mathbf x,y)$. However, this objective is
unsatisfactory, since it does not contain any supervised learning signal for $q_\phi(y | \mathbf x)$. Letting $\mathbf x_u$ denote unsupervised input variables and $(\mathbf x_s, y_s)$ denote supervised observation-latent pairs, \cite{kingma2014semi} propose to maximize the corresponding supervised and unsupervised objectives:
\begin{align}
\mathcal{L}^{\text{s}}(\mathbf x_s, y_s) &= \mathbb E_{q_\phi(\mathbf z_{}|y_s,\mathbf x_s)}\left[\log\left( \frac{p(\mathbf x_s, y_s, \mathbf z_{ })}{q_\phi(\mathbf z_{ }|y_s,\mathbf x_s)}\right)\right]\nonumber\\
 &+ \alpha~\mathbb E_{\hat p(y_s,\mathbf x_s)}\left[ \log(q_\phi(y_s|\mathbf x_s)) \right]
\label{eq:sup}
\end{align}
\begin{equation}
\small{\mathcal{L}^{\text{u}}(\mathbf x_u) = \mathbb E_{q_\phi(\mathbf z,y|\mathbf x_u)}\left[\log\left( \frac{p(\mathbf x_u, y, \mathbf z_{})}{q_\phi(\mathbf z_{},y|\mathbf x_u)}\right)\right]}
\end{equation}

The additional second term in Equation~\ref{eq:sup} is an expectation over the empirical distribution of labeled training pairs, $\hat p(\mathbf x_s, y_s)$, and can be optionally scaled by a hyperparameter, $\alpha$, controlling the relative strength of supervised and unsupervised learning signals. A semi-supervised VAE typically achieves performance superior to its unsupervised version. We subsequently refer to this objective as ``M1+M2''.

\subsection{VARIATIONAL RNN}

\begin{figure}
    \center
    \includegraphics[width=0.8\linewidth]{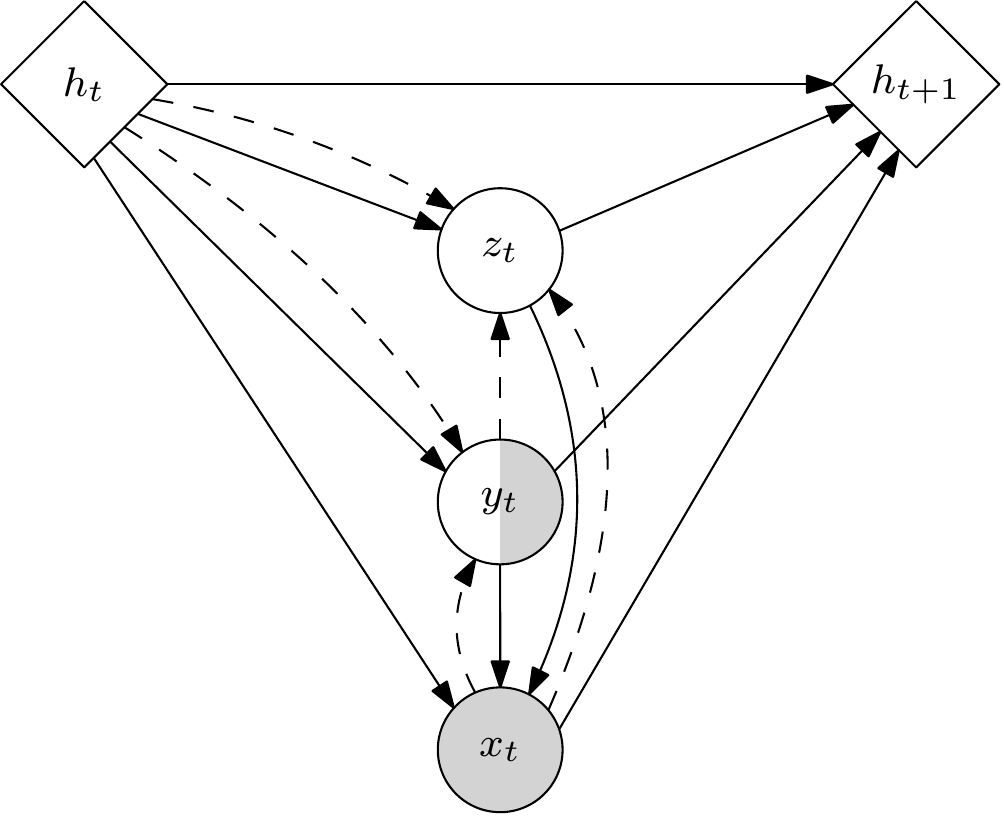}
    \caption{Per time step graphical model for the structured VRNN we use in this paper.
    In addition to the continuous latent variable $\mathbf z_{t}$, we include a discrete latent variable $y_{t}$ corresponding to an interpretable label, for which ground truth is sometimes available in the dataset.
    Dashed lines indicate the inference inference network, and solid lines the generative model.}
\label{fig:diag}
 \vskip 0.1in
\end{figure}

The variational RNN (VRNN) \citep{chung2015recurrent} is a deep latent generative model that is an extension of the VAE family. It can be viewed as an instantiation of a VAE at each time-step, with the model factorised in time overall. We use a structured VRNN variant, where a discrete latent variable $y$ is also present at each time step, as depicted in Figure~\ref{fig:diag}.

\subsection{RELATED WORK}

\cite{xu2017variational} introduce a semi-supervised VAE which overcomes the high-variance gradient estimators in sequential models using control variates during training. We build upon their work by adapting a similar architecture specification for the inference networks while avoiding the high-variance gradient estimators using reweighted wake-sleep style updates for training. Additionally, \cite{chen2018variational} proposes a semi-supervised DGM for modeling natural language using full sentence context. While a useful method for the task domain, they state their model cannot do generation.  

\section{CAFFEINATED WAKE-SLEEP}

Here we derive a general purpose training objective used for semi-supervised generative modeling. As we will show in subsequent experiments, both the objective and training method can be used to effectively train sequential models. As such, we introduce CWS in the context of time-series modeling, but note that it can trivially be used for static models by considering them to have a single time-step. In all subsequent sections, we denote $a_{\leq t} := a_{1:t}$ and $b_{< t} := b_{1:t - 1}$ if $t > 1$.

\subsection{GENERATIVE MODEL}

First, we define a generative model over a sequence of observed variables, $\mathbf{x}_{1:T}$, continuous latent variables, $\mathbf{z}_{1:T}$, and a discrete latent categorical variable, $y_{1:T}$.
\begin{align}
	p_\theta(\mathbf{x}_{\leq T}, &y_{\leq T}, \mathbf{z}_{\leq T}) = \prod_{t \leq T} \Big[ p_\theta(\mathbf{x}_{t} | \mathbf{z}_{\leq t}, y_{\leq t}) \label{eq:p} \\
	&\times p_{\theta}(\mathbf{z}_{t} | \mathbf{x}_{<t}, y_{<t}, \mathbf{z}_{<t})~p_{\theta}(y_{t} | \mathbf{x}_{<t}, y_{<t}, \mathbf{z}_{<t}) \Big]\nonumber 
\end{align}
Additionally, we are given $\mathbf x_{\leq T}$ and a subset $S \subseteq 1:T$ of labeled $y_t$, denoted $y_S := \{y_t: t \in S\}$.
We denote unlabeled $y_t$ as $y_U := \{y_t: t \in (1:T) \setminus S\}$.

  \begin{figure*}
    \includegraphics[width=\textwidth]{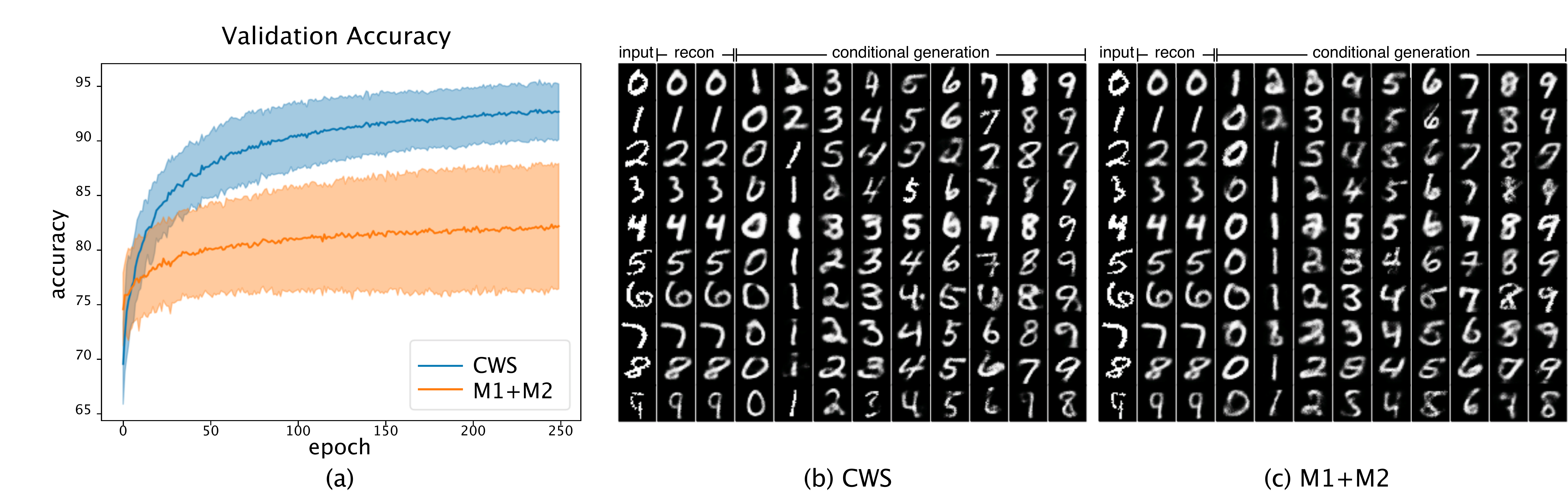}
    \caption{
(a) Validation accuracy of a MNIST classification model comparing two approaches for semi-supervised learning, CWS and M1+M2. As seen, CWS trains to the highest validation accuracy and faster on a held out test set of 10000 examples. (b) Visual reconstructions of MNIST digit in final trained model using CWS (c) Reconstructions using the final trained model by M1+M2 method. In (b-c), the first column denotes the input image, the second denotes reconstruction where $z,y$ are both inferred using $q(y,z|x)$, the third denotes reconstruction where $z$ is inferred given the correct label $y$, and columns 4-12 denote reconstruction where we infer the style, $z$, fix it, and vary the $y$ label to conditionally generate.}
\label{fig:mnist}
\vskip 0.1in
\end{figure*}

\subsection{INFERENCE}
Given the generative model defined above and a partially labeled dataset in the $y$ space, our variational distribution is the following:
\begin{align}
    &q_\phi(y_U, \mathbf{z}_{\leq T} \given \mathbf{x}_{\leq T}, y_S) = \label{eq:q}\\
    &~~~~~~\prod_{t \in U} q_\phi(y_t \given \mathbf{x}_{\leq T}, y_{<t}, \mathbf{z}_{< t}) \prod_{t \leq T} q_{\phi}(\mathbf{z}_t \given \mathbf{x}_{\leq T},y_{\leq t}, \mathbf{z}_{< t})\nonumber
    \end{align}
If the categorical variable $y_{t}$ is always treated as a latent variable in the fully unsupervised case, we can define a variational distribution, $q(\mathbf{z}_{\leq T}, y_{\leq T} | \mathbf{x}_{\leq T})$ and maximize the ELBO with respect to $\{\theta, \phi\}$. Instead, we will not need to infer given $y_S$ for any sequence, but at any time-step when $y_t$ is available, we use it instead of a sampled $y_t$ from the variational distribution. This is made clear by writing the \gls{ELBO} as:
\begin{align}
&\log~p({\mathbf{x}_{\leq T}, {y}_{S}}) \geq \\
&\small{\mathbb E_{q_\phi(y_U, \mathbf{z}_{\leq T} \given \mathbf{x}_{\leq T}, y_S)} \bigg[ \log \bigg( \prod_{t \in S}
\frac{p_\theta(\mathbf{x}_{t},\mathbf{z}_{t}, y_{t}, )}
	 {q_\phi(\mathbf{z}_t \given \mathbf{x}_{\leq T},y_{\leq t }, \mathbf{z}_{<t})}
} \nonumber \\
&~~~~~~~~~~~~~~~~~~~~~~~~~~~~~~\times\small{\prod_{t \in U}
\frac{p_\theta(\mathbf{x}_{t},\mathbf{z}_{t}, y_{t})}
	 {q_\phi(y_t, \mathbf{z}_{t} \given \mathbf{x}_{\leq T},\mathbf{z}_{<t}, y_{< t })}}\bigg)\bigg]\nonumber
\end{align}

\subsection{OPTIMIZATION}
Having defined the model and inference network, we now specify the optimization objective. 

\subsubsection{SEMI-SUPERVISED WAKE-SLEEP}
Before introducing our new objective, we first show how to use the reweighted wake-sleep algorithm \citep{bornschein2014reweighted} with the semi-supervised objective introduced by \cite{kingma2014semi}. Following the recommendation of \cite{le2018revisiting}, extending reweighted wake-sleep to include a supervision term can be used to effectively avoid problems with discrete variables. Below we derive the semi-supervised wake-sleep (SSWS) objective within that framework. Both of our objectives will be evaluated by sampling from $q_\phi$, so for each $k \in 1:K$ we have $y_t^k \sim q_\phi(y_t \given \cdot)$ for all $t \in U$ and $\mathbf{z}_t^k \sim q_\phi(\mathbf{z}_t \given \cdot)$ for all $t \in 1:T$ with a convention $y_t^k = y_t$ for $t \in S$.

For learning the generative model, we maximize the IWAE bound with respect to parameters $\theta$:
\begin{align}
    \mathcal L_\mathrm{p}(&\mathbf{x}_{\leq T}, y_S) := \nonumber \\
    &\E_{q}\left[\log \frac{1}{K} \sum_{k = 1}^K \frac{p_\theta(y_U^k, \mathbf{z}_{\leq T}^k, \mathbf{x}_{\leq T}, y_S)}{q_\phi(y_U^k, \mathbf{z}_{\leq T}^k \given \mathbf{x}_{\leq T}, y_S)}\right],\label{eq:p-loss}\\
    &q = \prod_{k = 1}^K q_\phi(y_U^k, \mathbf{z}_{\leq T}^k \given \mathbf{x}_{\leq T}, y_S) \nonumber  
\end{align}
This is a lower bound to $\log p_\theta(\mathbf{x}_{\leq T}, y_S)$ which is tight when $q_\phi(y_U, \mathbf{z}_{\leq T} \given \mathbf{x}_{\leq T}, y_S) = p_\theta(y_U, \mathbf{z}_{\leq T} \given \mathbf{x}_{\leq T}, y_S)$.
Sampling from $q_\phi(y_U, \mathbf{z}_{\leq T} \given \mathbf{x}_{\leq T}, y_S)$ and evaluating its density is simple since it is factorized as in \eqref{eq:q} where both $q(y_t \given \cdot)$ and $q_\phi(\mathbf{z}_{t} \given \cdot)$ are given, and all values to the right of the conditioning bar are always available at sampling at time step $t$ (previous $y_{<t}$ are either sampled or given supervision and $\mathbf{z}_{<t}$ are sampled).

\begin{figure*}[tb]
  	\begin{center}
  		\includegraphics[width=\textwidth]{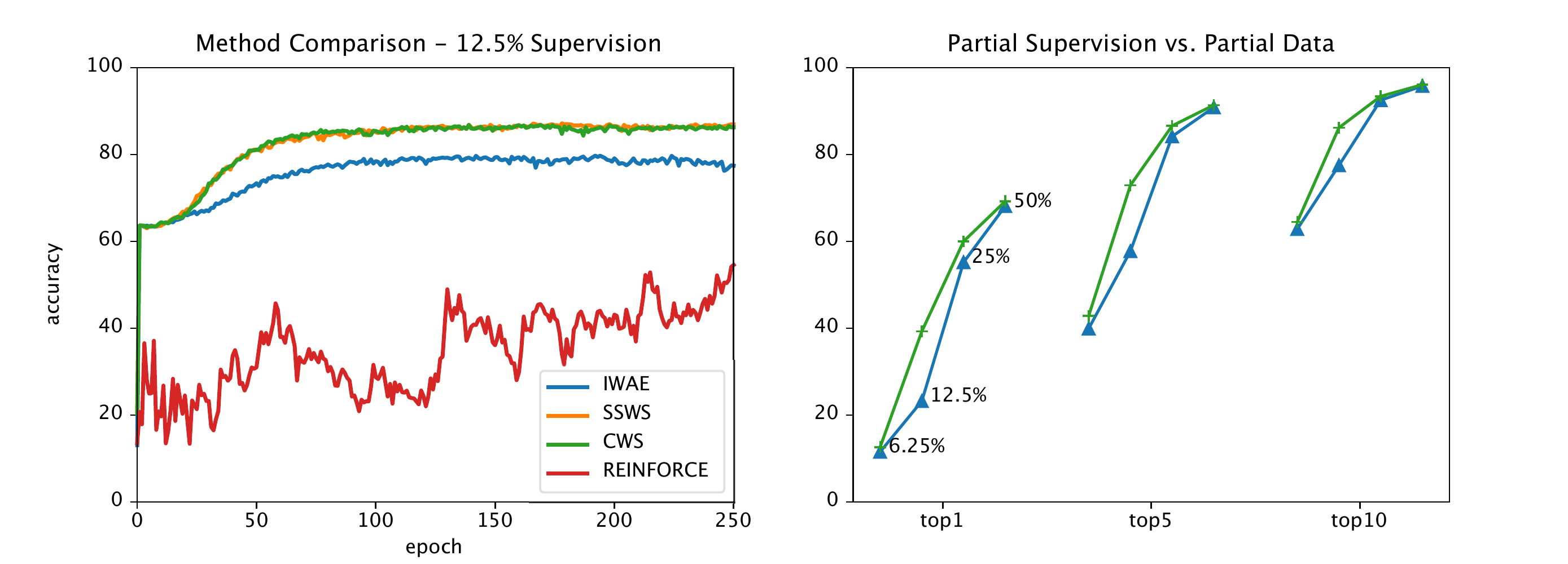}
  		\caption{ Left: Top10 validation accuracy throughout training for a 70-class character classification task comparing three methods of semi-supervised training, CWS, SSWS, and REINFORCE (M1+M2), with 12.5\% supervision rate and one method of fully supervised training using IWAE and a classifier loss on 12.5\% of the dataset. Right: Final top1, top5, and top10 accuracy summaries for semi-supervision using CWS (green) against models trained with full supervision but on a corresponding fraction of the original dataset (blue). As before, the method denoted by blue trains with full supervision using IWAE and a classifier loss on the subset of the total data denoted by the labels. }\label{hw-accs}
  	\end{center}
  	\vskip 0.1in
 \end{figure*}
 
For learning the inference network parameters, we need to consider the unsupervised and the supervised cases. For unsupervised inference learning, we minimize the expected $\KL$-divergence between the true posterior and the variational posterior given by $q_\phi$ under the generative model, $p := p_\theta(\mathbf{x}_{\leq T},y_S)$:
{\small \begin{align}
    &\nabla_\phi\E_{p}[\small{\KL(p_\theta(y_U, \mathbf{z}_{\leq T} \given \mathbf{x}_{\leq T}, y_S) || q_\phi(y_U, \mathbf{z}_{\leq T} \given \mathbf{x}_{\leq T}, y_S))}] \nonumber \\
    &= \E_{p}[\E_{p_\theta(y_U, \mathbf{z}_{\leq T} \given \mathbf{x}_{\leq T}, y_S)}\left[-\nabla_\phi\log q_\phi(y_U, \mathbf{z}_{\leq T} \given \mathbf{x}_{\leq T}, y_S)\right]] \nonumber
\end{align}}
Given some $(\mathbf{x}_{\leq T}, y_S)$ from the data distribution, $\hat p_\theta(\mathbf x_{\leq T},y_S)$, the inner expectation is approximated using samples from the inference network, $q_\phi$, referred to as the wake-$\phi$ update for learning parameters $\phi$:
\begin{align}
	    &\mathcal L_\mathrm{q}(\mathbf x_{\leq T}, y_S) := \nonumber \\
	    &\E_{q_\phi}\left[\sum_{k = 1}^K \bar w_k \left(-\log q_\phi(y_U^k, \mathbf{z}_{\leq T}^k \given \mathbf{x}_{\leq T}, y_S) \right)\right], \label{eq:q-loss}
\end{align}
where $\bar w_k$ is a normalized version of
\begin{align}
    w_k := \frac{p_\theta(y_U^k, \mathbf{z}_{\leq T}^k, \mathbf{x}_{\leq T}, y_S)}{q_\phi(y_U^k, \mathbf{z}_{\leq T}^k \given \mathbf{x}_{\leq T}, y_S)},~~~\bar w_k = \frac{w_k}{\sum_l w_l}\label{eq:weight}
\end{align}
and
\begin{equation}
q_\phi = \prod_k q_\phi(y_U^k, \mathbf{z}_{\leq T}^k \given \mathbf{x}_{\leq T}, y_S) \nonumber
\end{equation}
Evaluating \eqref{eq:q-loss} requires evaluating the joint $p$ density given in \eqref{eq:p}, the $q$ density given in \eqref{eq:q}, and sampling again from \eqref{eq:q}.

Finally, we need to include a supervised loss term. There is only supervision signal for some given $y_t$ for some $t \in S$ per data example. As such, we want to maximize the log-likelihood of the supervised pairs $(y_S, \mathbf{x}_{\leq T})$ under the variational distribution, $q_\phi(y_{\leq T}|\mathbf{x}_{\leq T})$, marginalized over unsupervised time-steps:
\begin{align}
    &q_\phi(y_{S}|\mathbf{x}_{\leq T}) = \int q_\phi(y_{\leq T}, \mathbf{z}_{\leq T} \given \mathbf{x}_{\leq T}) dy_U d\mathbf z_{\leq T} \nonumber \\
    &= \int q_\phi(y_{S}|\mathbf{x}_{\leq T}, \mathbf{z}_{\leq T}, y_U )q_\phi(y_{U}, \mathbf{z}_{\leq T} \given \mathbf{x}_{\leq T}) dy_U d\mathbf z_{\leq T} \nonumber \\
      &= \E_{q_\phi(y_{U}, \mathbf{z}_{\leq T} \given \mathbf{x}_{\leq T})}[q_\phi(y_{S}|\mathbf{x}_{\leq T}, \mathbf{z}_{\leq T}, y_U )]
\end{align}

We lower bound this with Jensen's inequality and obtain our supervised loss term:
\begin{align}
\mathcal L_{\mathrm{s}}(& y_S) := \nonumber \\ 
&\E_{q_\phi(y_{U}, \mathbf{z}_{\leq T} \given \mathbf{x}_{\leq T})}[\log q_\phi(y_{S}|\mathbf{x}_{\leq T}, \mathbf{z}_{\leq T}, y_U )] \label{eq:s-loss}
\end{align}
For each minibatch the objectives, \eqref{eq:p-loss}, \eqref{eq:q-loss}, and \eqref{eq:s-loss}, are computed using the same set of samples
and we perform the relevant gradient steps in $\phi$ and $\theta$ by alternating between the two parameter updates:
\begin{align}
&\theta^{*} = \theta + \alpha_\theta \nabla_\theta  \mathcal L_\mathrm{p}(\mathbf x_{\leq T}, y_S)	\\
\phi^{*} &= \phi - \alpha_\phi \nabla_\phi \Big(\mathcal L_\mathrm{q}(\mathbf x_{\leq T}, y_S) - \mathcal L_{\mathrm{s}}(y_S)\Big)
\end{align}

\begin{figure*}[tb]
  	\begin{center}
  		\includegraphics[width=\textwidth]{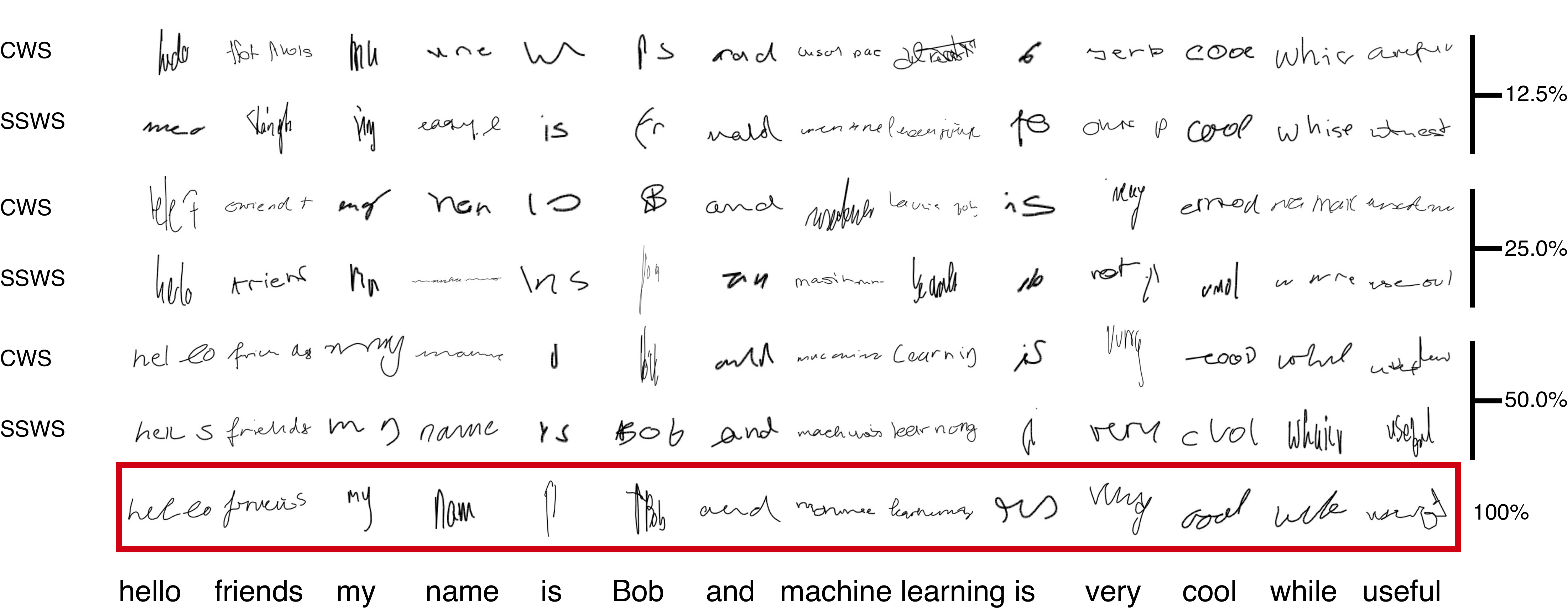} 
  		\caption{Generated of sample trajectories of handwriting conditioned on generating the bottom text. Each pair of rows is trained with a supervision
rate (right) and a method (left). Corresponding sample generated from a fully supervised model is displayed in the red box. Note that the 100\% supervision example in this figure is trained with the full dataset with a separate classifier loss.} \label{visual-recon}
  	\end{center}
  	\vskip 0.1in
 \end{figure*}
 
\subsubsection{CAFFEINATED WAKE-SLEEP}
Although the SSWS approach can be used to train time-series models, in practice there is a tradeoff between learning and optimization stability. As mentioned earlier, sequences of observations may contain variable amounts of supervision, the most extreme example being datasets containing both sequences with fully observed and fully unobserved labels.

Because of this, the magnitude of the supervised and unsupervised terms in the $q_\phi$ loss will vary per data sequence, and we incur a tradeoff between correcting for this bias and computational efficiency. For example, we could normalize $\mathcal L_{\mathrm{s}}$ and $\mathcal L_{\mathrm{q}}$ by $|S|$ and $|U|$, respectively, but doing so treats sequences with a lower supervision rate the same as those with much higher supervision. Alternatively, we could take a weighted average of the terms across the sequences, but this faces the same issue as before unless we also scale the learning rate dynamically per gradient step for each stochastic mini-batch.

To remedy this problem, we obviate the need to maintain two different supervision terms by minimizing the expected $\KL$-divergence between the true posterior and the full variational posterior under the generative likelihood, $p_\theta(\mathbf{x}_{\leq T})$:
{\small\begin{align}
    &\nabla_\phi \E_{ p_\theta(\mathbf{x}_{\leq T})}[\small{\KL(p_\theta(y_{\leq T}, \mathbf{z}_{\leq T} \given \mathbf{x}_{\leq T}) || q_\phi(y_{\leq T}, \mathbf{z}_{\leq T} \given \mathbf{x}_{\leq T}))}] \nonumber\\
     &= \E_{ p_\theta(\mathbf{x}_{\leq T})}[\E_{p_\theta(y_U, y_S, \mathbf{z}_{\leq T} \given \mathbf{x}_{\leq T})}\left[-\nabla_\phi \log q_\phi(\cdot \given \mathbf{x}_{\leq T})\right]] \nonumber
\end{align}}

Unlike in the derivation for Equation~\eqref{eq:q-loss}, there is no distinction here between $y_S$ and $y_U$, meaning that we should always sample a value for $y_t$ when computing this loss. However, doing so would give the fully unsupervised wake-$\phi$ objective and not include any supervision signal for the $y_S$ labels that we do have. To ``correct'' for this, we introduce an empirically justified bias into this estimator, namely replacing sampled values of $y_t$ with $y_S$ when available. Intuitively, this is exactly the reweighted wake-sleep algorithm if $y_S$ is treated as having been sampled from the distributions, $q_\phi(\cdot)$. This is a kind of teacher-forcing approach \citep{williams1989learning} for discrete latent labels, $y_t$.

To be concrete about our method, we still compute an expectation over the empirical data distribution $\hat p(\mathbf{x}_{\leq T}, y_S)$, and estimate the innermost expectation with samples from \eqref{eq:q} when needed. The bias introduced then comes from the additional $q_\phi(y_S|...)$ terms in the denominator of the importance weights:
\begin{align}
	    &\mathcal L^\mathrm{CWS}_{\mathrm{q}}(\mathbf x_{\leq T}, y_S) := \nonumber \\
	    &\E_{(y_S, \mathbf{x}_{\leq T})\sim \hat p, ~~(y_U^{1:K}, \mathbf{z}^{1:K})\sim q_\phi(y_U^k, \mathbf{z}_{\leq T}^k \given y_S, \mathbf{x}_{\leq T})}\left[f\right], \nonumber \\
	    &f := \sum_{k = 1}^K \tilde w_k \left(-\log q_\phi(y_U^k, y_S, \mathbf{z}_{\leq T}^k \given \mathbf{x}_{\leq T}) \right), \label{eq:cws-loss}\\
	    &w_k^{\mathrm{cws}} := \frac{p_\theta(y_U^k, \mathbf{z}_{\leq T}^k, \mathbf{x}_{\leq T}, y_S)}{q_\phi(y_U^k, y_S, \mathbf{z}_{\leq T}^k \given \mathbf{x}_{\leq T})},~~~\tilde w_k = \frac{w_k^{\mathrm{cws}}}{\sum_l w_l^{\mathrm{cws}}}\nonumber
\end{align}
Finally, we formally introduce the caffeinated wake-sleep algorithm. Model parameters are learned with the importance-weighted \gls{ELBO} as before. However, crucially, we use a unified gradient estimator in \eqref{eq:cws-loss} to update $\phi$ parameters:
\begin{align}
&\theta^{*} = \theta + \alpha_\theta \nabla_\theta  \mathcal L_\mathrm{p}(\mathbf x_{\leq T}, y_S)	\\
\phi^{*} &= \phi - \alpha_\phi \nabla_\phi \mathcal L_\mathrm{q}^{\mathrm{CWS}}(\mathbf x_{\leq T}, y_S)
\end{align}


\begin{figure*}[tb]
\label{fig:flies}
  	\begin{center}
  		\includegraphics[width=0.9\textwidth]{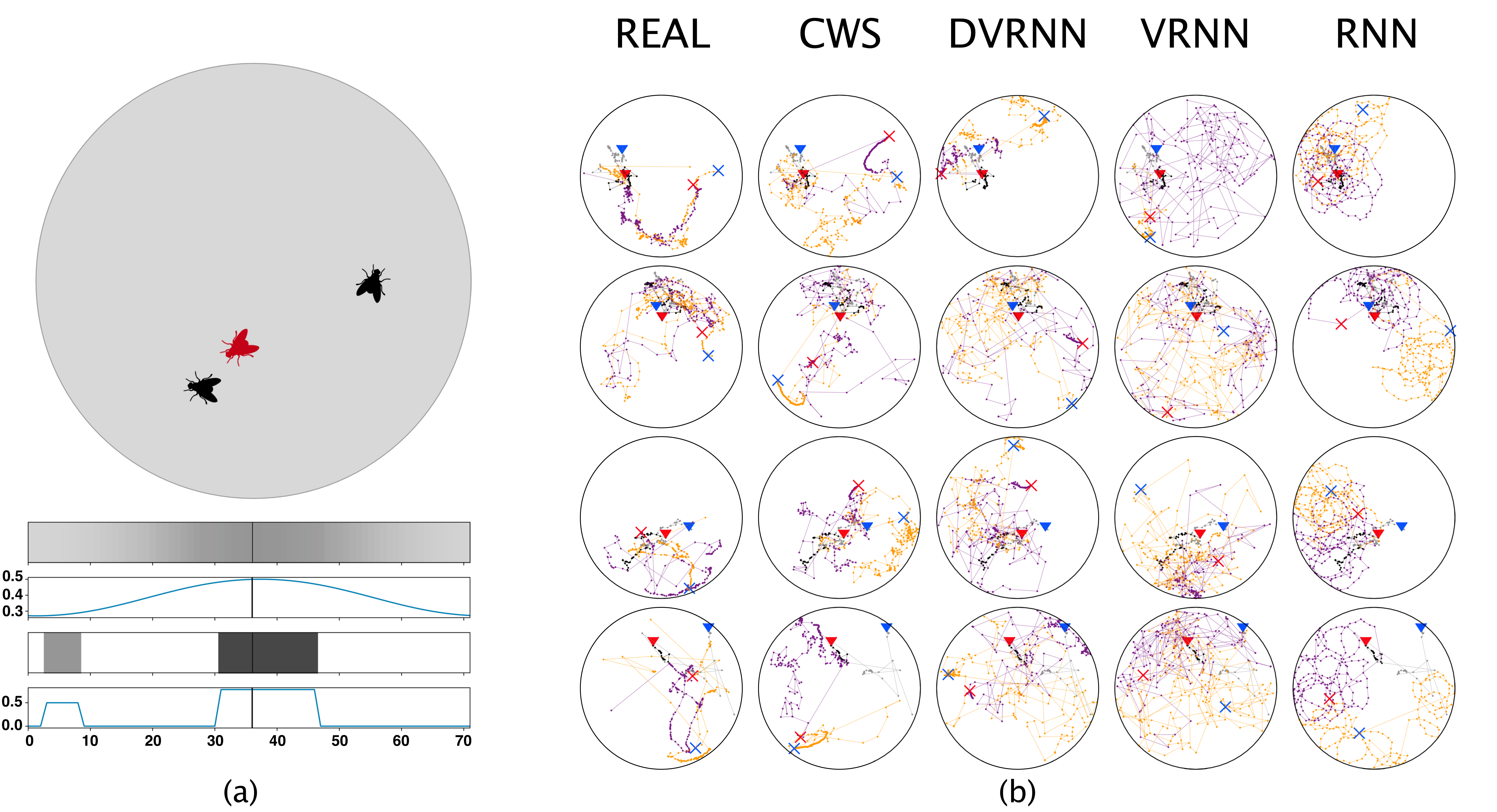}
  		\caption{(a) Diagram of the red fly's field of view encoding in a petri dish environment containing two other flies. The middle line in the plots denotes the fly's direct line of sight. Each pair of plots indicates the agent's field of view with respect to walls and other flies, respectively. The fly closer to the red agent contributes more mass to the encoding vector. (b) Comparison of each model's continuation tracks for four distinct seed sequences. Each column shows a model's continuation sampled from the generative model compared against the leftmost column displaying the ground truth continuation. Each row shows one of four seed sequences used to seed the generative models. Within the petri dish, both flies' locomotion tracks are shown. For the first fly, the red arrow indicates the starting position of the seed sequence, the black markers and line indicate the seed tracks, the purple markers and line indicate the sampled continuation tracks, and the red 'x' indicates the final position of the fly after 200 time-steps. For the second fly, these indicators are blue arrow, gray markers, orange markers, and blue 'x', respectively.}

  		\label{fig:flies}
  	\end{center}
  	\vskip 0.1in
 \end{figure*}

\begin{figure*}[tb]
  	\begin{center}
  		\centerline{\includegraphics[width=0.95\textwidth]{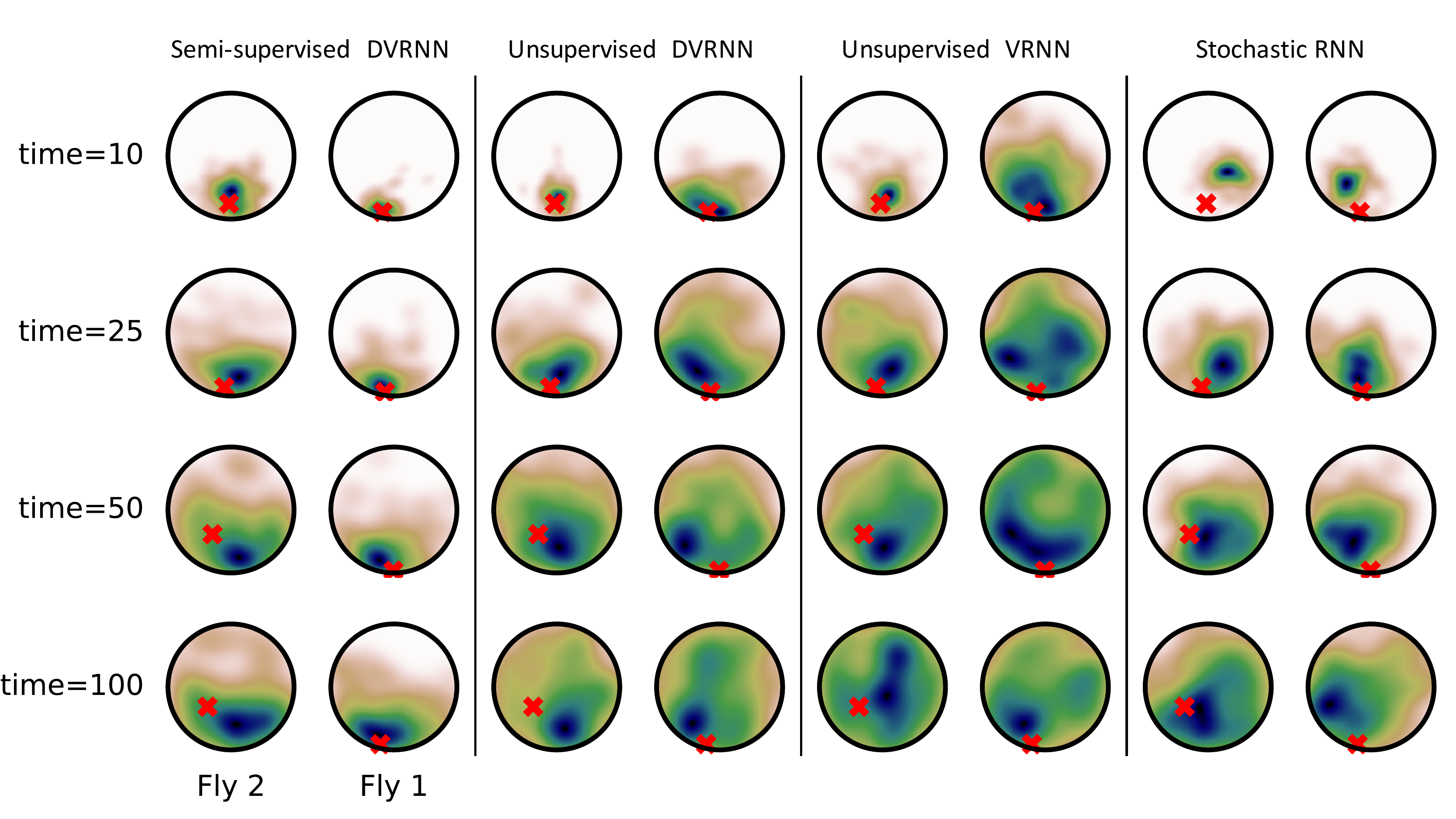}}

  		\caption{ Comparison of uncertainty estimates over future trajectories of two flies' interactions in the environment between different models. Each model is seeded with the same ground truth sequence each time and 100 continuations are sampled for 100 time-steps into the future. We show the kernel density estimate of all fly positions at the indicated time-step across the 100 continuations. Each model's estimate is split by fly for clarity (two columns per model indicated by separators), where the red 'x' indicates the true position of the fly at that time-step. The future trajectories across all models show greater uncertainty about the fly's position as the model evolves in time. We find the model trained with CWS is the least noisy as it evolves, still captures the actual position within a high probability region, and has the most mass close to the true position of fly 1 at time 100 among the 4 models. }
      \label{fig:uncertainty}
  	\end{center}
  	\vskip 0.1in
  \end{figure*}

\section{EXPERIMENTS}

We evaluate CWS on an variety of tasks within generative modeling by running experiments on three separate datasets, MNIST, IAM On-Line Handwriting Database (IAM-OnDB) \citep{liwicki2005iam}, and Fly-vs-Fly \citep{eyjolfsdottir2014detecting}. In our experiments, we consider the training accuracy of the inference network in classification, the conditional and unconditional generation of new data, and the uncertainty captured by our generative models. 

\subsection{SEMI-SUPERVISED MNIST}

We start with a toy MNIST experiment, which semi-supervises a discrete latent variable corresponding to digit class. We use the same model and VAE architecture detailed in \cite{kingma2014semi} and compare against their training method, denoted as M1+M2 with the supervised weight set to 60. For supervision, we use only 100 labeled digits out of a total dataset of 50000. We briefly note that CWS for the single time-step case is trivial given the objectives defined above and SSWS recovers the M1+M2 objective.

In Figure~\ref{fig:mnist}a, we compare validation accuracy of digit classification between CWS and M1+M2 trained models. We find that our method both trains faster and better despite using an identical generative model, inference network, learning rate, and optimizer. Furthermore, our approach requires setting one fewer hyperparameter when compared to M1+M2.

In Figure~\ref{fig:mnist}b, we conditionally generate samples using our trained models by first inferring the style using $q_\phi(\mathbf{z}|\mathbf{\hat x},\hat y)$.  Then, fixing the style, we change $y$ and reconstruct using $p_\phi(\mathbf{x}|\mathbf{z},y)$. We find that CWS is able to conditionally generate as well as prior art.

\subsection{HANDWRITING}

Next, we run an experiment on the IAM-OnDB dataset. The generative model is over $\mathbf{x}_t := \{c^i_t, c^j_t, \mathrm{pen}_t, \mathrm{eoc}_t\}$, where $c_i, c_j$ denotes a single time-step vectorized stroke of the pen and the $\mathrm{pen}_t$ and $\mathrm{eoc}_t$ denote pen-up status and end-of-character binary values. The latent space in this model is comprised of continuous style variable, $\mathbf{z}_t$, and a discrete character label corresponding to a sequence of observed strokes that make up a valid character, $y_t$. For the IAM-OnDB data, the valid alphabet is comprised of 70 different characters: upper and lower case letters, digits, and special characters. The large latent space makes this a very challenging time-series problem.

The VRNN architecture we use is taken from the DeepWriting model introduced by \cite{aksan2018deepwriting}. At a high level, we use a VRNN over $\mathbf{x}_{\leq T}, \mathbf{z}_{\leq T}$, and a BiLSTM network for $q_\phi(y_{\leq T}|\mathbf{x}_{\leq T})$. Because the VRNN can be optimized solely using reparameterization, we train $q_\phi(\mathbf{z}_{\leq T}| y_{\leq T}, \mathbf{x}_{\leq T})$ using IWAE and $q_\phi(y_{\leq T}|\mathbf{x}_{\leq T})$ using CWS. This mirrors the optimization in \cite{aksan2018deepwriting}, except that now, we can train with semi-supervision. For more detailed experimental setup, we refer the reader to the cited work. In all experiments, we use a training set of 26560 sequences and a validation set of 512 sequences, all of length 200. 

As previously mentioned, techniques for semi-supervised learning do not trivially extend to time-series models. Instead, we compare CWS against SSWS, the latter of which can be viewed as the M1+M2 objective for sequential models using wake-$\phi$ style updates for learning discrete latents. For comparison, we can also train the M1+M2 objective without using RWS by using the REINFORCE estimator for taking gradients through discrete latent variables. In Figure~\ref{hw-accs} (left), we find that this method results in poorly trained inference networks which cannot classify character labels accurately even using top-10 metric.

In Figure~\ref{hw-accs} (right), we present classification results of training
with a fixed architecture, varying only the training objective using either CWS or IWAE. Because the normal
IWAE objective cannot be used with semi-supervision
and REINFORCE, we compare CWS with fully unsupervised IWAE plus training the classifier separately with
full supervision. This baseline is generated by using the
optimization technique used from \cite{aksan2018deepwriting} on
a corresponding fraction of the dataset. In other words,
we take the exact same architecture and compare training with conventional DGM techniques plus a classification loss against training with our CWS method using
the same amount of full labels but being able to incorporate unlabeled data. We find that validation accuracy
of the classifier trained with CWS using additional unlabeled data greatly outperforms the IWAE baseline using
the same amount of partial labels from the dataset but
without the additional data.

In Figure~\ref{visual-recon}, we show reconstructed samples from trained
models where we force the network to attempt to generate the phrase, "hello friends my name is Bob and machine learning is very cool while useful". We find that
even at 12.5\% supervision, sampled generations resemble the target sentence using SSWS or CWS. 

  \subsection{FLY TRACKING}  
In our final experiment, we run CWS on a dataset of two male fruit flies interacting in a petri dish \citep{eyjolfsdottir2014detecting}. We use a VRNN which attempts to mirror biological plausibility by modeling movement, $\mathbf{x}_t$, as velocities of body position and configuration \citep{cueva2018emergence}. We inject as input to the RNN at each time-step, the visual encoding for a given fly shown in Figure~\ref{fig:flies}a \citep{clandinin2015neuroscience} along with knowledge of its own state \citep{cueva2018emergence}.

The dataset uses high level actions, which we model as a 6-valued discrete variable, $y_{t}$, corresponding to possible semantic labels:
"lunge", "charge", "tussle", "wing threat", "hold", and "unknown". These annotations
are provided by human experts at time-steps dispersed throughout the dataset
without a clear pattern or regularity. In all experiments we compare four models: the RNN used by
\cite{eyjolfsdottir2017learning} outputting a probability vector over
discretized action space at each time step, the standard VRNN without $y$ and
with continuous valued $\mathbf{z}$, the discrete VRNN (DVRNN) with $y_t$ trained unsupervised, and the same DVRNN trained with CWS. While CWS was able to train this model, we ran into optimization difficulties with the same model but training with a separate supervision term. We fine-tune and report results of the best performing model during training: CWS-DVRNN with $30$-dimensional $\mathbf{z}$ and $6$-valued $y$, RWS-DVRNN with $50$-dimensional $\mathbf{z}$ and $10$-valued $y$, and VRNN with $120$-dimension $\mathbf{z}$. We provide further details of model in the Appendix. 

In Figure~\ref{fig:flies}b, we condition the model on an initial sequence of
actions, then sample a continuation and visually inspect how it compares with
the true continuation that was not shown to the model. We find that under the RNN model, the flies tend to
move in circular patterns with relatively constant velocity. In contrast, real
flies tend to alternate between fast and slow movements, changing their directions
much more abruptly. All three variants of our VRNN model qualitatively recover
this behavior, however the continuous VRNN also tends to behave too erratically. In the DVRNN models, we were not able to identify any other clear visual
artifacts in the generated trajectories that disagrees with the real data,
although DVRNN without semi-supervision appears to generate trajectories where flies move too much.
\begin{table}
\vskip 0.2in
\caption{KDE of ground truth position under model} \label{kde}
\begin{center}
  \begin{tabular}{lll}
    \toprule
{}&\multicolumn{2}{c}{KDE $\log p$}\\
\cmidrule(r){2-3}
\multicolumn{1}{c}{\bf model}&\multicolumn{1}{c}{\bf fly 1}&\multicolumn{1}{c}{\bf fly 2}\\
\hline \\
\multicolumn{1}{c}{CWS+DVRNN}  &\multicolumn{1}{c}{$\bm{ -780.1}$} & \multicolumn{1}{c}{$\bm{-851.5}$} \\
\multicolumn{1}{c}{RWS+DVRNN} & \multicolumn{1}{c}{$-901.8$} & \multicolumn{1}{c}{$-853.9$}\\
\multicolumn{1}{c}{VRNN}  & \multicolumn{1}{c}{$-962.0$} & \multicolumn{1}{c}{$-879.8$} \\
\multicolumn{1}{c}{RNN} & \multicolumn{1}{c}{$-931.0$} & \multicolumn{1}{c}{$-921.2$}\\
    \bottomrule
  \end{tabular}
   \end{center}
   \vskip -0.1in
  \end{table}
In Figure~\ref{fig:uncertainty}, we investigate the quality of uncertainty estimates produced by various models. For this purpose we again seed the model with an initial sequence of actions, then
observe how the probability mass over the flies' projected future positions evolves over time and compare it with the actual positions in the dataset. Figure \ref{fig:uncertainty} visualizes the results and Table~\ref{kde} provides the log likelihood of each fly's true position under this density estimate. Again, we find that CWS trained DVRNN performs best, followed by the unsupervised DVRNN, the continuous VRNN, and the RNN.

\subsection{DISCUSSION}

We have introduced a new method for semi-supervised learning in deep generative time-series models and we have shown that it achieves better performance than unsupervised learning and fully supervised learning with a fraction of the data. Although CWS uses a biased gradient estimator, it has many advantages including ease of implementation, intuition as teacher-forcing RWS, and empirical validation. A formal, theoretical justification of why the biased estimator of CWS works remains for future work. Additionally, we are interested in extending CWS to a more general class of semi-supervised models, taking inspiration from the ideas of \cite{narayanaswamy2017learning}. 

\paragraph{Acknowledgements}~\newline
~\newline
Michael Teng is supported under DARPA D3M (Contract No.FA8750-19-2-0222)
and partially supported by the DARPA Learning with Less Labels (LwLL) program (Contract No.FA8750-19-C-0515). Tuan Anh Le is support by the AFOSR award (Contract No.FA9550-18-S-0003). We acknowledge the support of the Natural Sciences and Engineering Research Council of Canada (NSERC), the Canada CIFAR AI Chairs Program, and the Intel Parallel Computing Centers program.

\bibliography{cws}
\bibliographystyle{plainnat}
\setcitestyle{numbers}

\end{document}